\documentclass[sigconf]{acmart}
\AtBeginDocument{%
  \providecommand\BibTeX{{%
    \normalfont B\kern-0.5em{\scshape i\kern-0.25em b}\kern-0.8em\TeX}}}

\usepackage{subcaption}
\usepackage{algorithm}
\usepackage{algpseudocode}
\usepackage{hyperref}
\usepackage{cleveref}
\usepackage{tikz}
\usepackage[mathscr]{eucal}
\usepackage{amsbsy}
\usepackage{bm}
\usepackage{subcaption} 
\usepackage{pifont}
\usepackage{enumitem}

\usepackage{amsmath,amsfonts,bm}

\def\eqref#1{equation~\ref{#1}}

\def\1{\bm{1}}

\def\vg{{\bm{g}}}

\def\vx{{\bm{x}}}

\def\mG{{\bm{G}}}

\def\mP{{\bm{P}}}

\def\mW{{\bm{W}}}
\def\mX{{\bm{X}}}

\DeclareMathAlphabet{\mathsfit}{\encodingdefault}{\sfdefault}{m}{sl}
\SetMathAlphabet{\mathsfit}{bold}{\encodingdefault}{\sfdefault}{bx}{n}

\usepackage{pythonhighlight}

\newcommand{\eg}{e.g., }

\newcommand{\Tra}{{\sf T}} 
\newcommand{\M}[2][]{{\bm{#1\mathbf{\MakeUppercase{#2}}}}} 
\newcommand{\V}[1]{\bm{#1}} 

\newcommand{\update}[1]{\textcolor{black}{#1}}

\newcommand\encircle[1]{%
  \tikz[baseline=(X.base)] 
    \node[circle,fill=black,text=white,font=\sffamily\bfseries,inner sep=0.04cm] (X) {#1};}

\newcommand{\cmark}{\ding{51}}%
\newcommand{\xmark}{\ding{55}}%

\usepackage{xspace}
\newcommand{\mf}{Mosaic Flow\xspace}

\copyrightyear{2023} 
\acmYear{2023} 
\setcopyright{rightsretained} 
\acmConference[SC '23]{The International Conference for High Performance Computing, Networking, Storage and Analysis}{November 12--17, 2023}{Denver, CO, USA}
\acmBooktitle{The International Conference for High Performance Computing, Networking, Storage and Analysis (SC '23), November 12--17, 2023, Denver, CO, USA}
\acmDOI{10.1145/3581784.3613217}
\acmISBN{979-8-4007-0109-2/23/11}

\begin{document}

\title{Breaking Boundaries: Distributed Domain Decomposition with Scalable Physics-Informed Neural PDE Solvers}

\author{Arthur Feeney}
\authornote{Both authors contributed equally to this research.}
\email{afeeney@uci.edu}
\author{Zitong Li}
\authornotemark[1]
\email{zitongl5@uci.edu}
\affiliation{%
  \institution{University of California, Irvine}
   \country{USA}
}

\author{Ramin Bostanabad}
\email{raminb@uci.edu}
\affiliation{%
  \institution{University of California, Irvine}
   \country{USA}
}

\author{Aparna Chandramowlishwaran}
\email{amowli@uci.edu}
\affiliation{%
  \institution{University of California, Irvine}
   \country{USA}
}


\begin{abstract}
\mf is a novel domain decomposition method designed to scale physics-informed neural PDE solvers to large domains. Its unique approach leverages pre-trained networks on small domains to solve partial differential equations on large domains purely through inference, resulting in high reusability. \update{This paper presents an end-to-end parallelization of \mf, combining data parallel training and domain parallelism for inference on large-scale problems.} By optimizing the network architecture and data parallel training, we significantly reduce the training time for learning the Laplacian operator to minutes on 32 GPUs. Moreover, our distributed domain decomposition algorithm enables scalable inferences for solving the Laplace equation on domains $4096\times$ larger than the training domain, demonstrating strong scaling while maintaining accuracy on 32 GPUs. The reusability of \mf, combined with the improved performance achieved through the distributed-memory algorithms, makes it a promising tool for modeling complex physical phenomena and accelerating scientific discovery.
\end{abstract}



\maketitle

\section{Introduction}
Scientific machine learning (SciML) is an emerging field that aims to integrate scientific knowledge into the development of machine learning models. By leveraging domain expertise, SciML reduces the reliance on massive datasets that are often scarce or difficult to create in many scientific fields, such as fluid dynamics \cite{brunton2020machine}.
To achieve this integration, researchers have proposed various innovative strategies, ranging from incorporating scientific principles into deep neural network architectures and loss functions, developing hybrid models, using transfer learning and domain adaptation techniques, and employing Bayesian techniques \cite{karniadakis2021physics, markidis2021old, obiols2020cfdnet, obiols2021surfnet}.

Among the above approaches, physics-informed neural networks (PINNs) have shown promise for solving complex problems that involve partial differential equations (PDEs) by incorporating physical laws and constraints \cite{raissi2019physics}. By softly enforcing PDE constraints in the loss function, PINNs can learn from limited data and still provide accurate predictions. Unlike traditional methods, PINNs are mesh-free and time-continuous, making them attractive for many complex scientific applications. The growing interest in physics-informed machine learning has led to the development of numerous software libraries that offer an easy and efficient way to create PINNs. Some of the popular libraries include DeepXDE \cite{lu2021deepxde}, NVIDIA Modulus \cite{hennigh2021nvidia}, and SciANN \cite{haghighat2021sciann}.

While PINNs have shown great promise in solving problems in small domains with simple geometries, they face challenges when applied to larger domains. As the domain size increases, the complexity of the problem also grows, necessitating larger networks to capture the underlying features accurately.
Since the PINN loss function can be highly non-convex, larger networks can result in a stiff and hard optimization problem, leading to significantly slower convergence with reduced accuracy or no convergence at all \cite{krishnapriyan2021characterizing, wang2021understanding}. Additionally, training PINNs for large domains require significant computational resources, which can limit their applicability to real-world problems.

\begin{table*} [ht]
 \centering
 \caption{State-of-the-art domain decomposition methods for neural PDE solvers. The overlapping approaches are based on Schwarz methods or inspired by them as denoted by $\ast$. Unlike MosaicFlows which solves PDEs on arbitrarily large domains using only neural network inferences, all other approaches require training a new model for each new domain.} 
 \begin{tabular}{|r|l|l|l|l|c|c|}
		\hline
		DDM             &  Subdomains & PINN formulation & Interface & Interface & \update{Dist} & \update{Dist} \\
                        &             &          &   condition      & resolved   &   \update{alg} & \update{eval} \\\hline
        DPINN \cite{dwivedi2021distributed}   & non-overlapping & residuals   & loss terms & training & \xmark & \xmark \\
		XPINN \cite{jagtap2021extended,shukla2021parallel}                 & non-overlapping  & residuals  & loss terms &  training   & \cmark  &  \cmark\\ 
        cPINN \cite{jagtap2020conservative, shukla2021parallel}  & non-overlapping  & conservation laws     & loss terms & training & \cmark & \cmark\\
  		hp-VPINNs \cite{kharazmi2021hp} & non-overlapping  & variational residuals   & loss terms    & training & \xmark  & \xmark  \\ 
        DeepDDM \cite{li2020deep}  & overlapping  & residuals          & Schwarz & training & \xmark & \xmark\\
        D3M \cite{li2019d3m}     & overlapping  &    variational residuals    & Schwarz & training & \xmark & \xmark\\
		FBPINN \cite{moseley2021finite, dolean2022finite}           & overlapping    & residuals    & Schwarz$^{\ast}$    & training & \cmark & \xmark  \\ 
  		\mf \cite{wang2022mosaic}        & overlapping     & residuals       & Schwarz$^{\ast}$  & inference & \xmark  & \xmark \\
        \textbf{Dist-MF (this paper)}        & overlapping     & residuals       & relaxed Schwarz  & inference & \cmark  & \cmark \\
		\hline
\end{tabular}
  \label{tab:ddm}
\end{table*}

Domain decomposition \cite{dolean2015introduction} has emerged as a potential solution to improve the scalability and convergence of neural PDE solvers on large domains. This approach involves breaking down the challenging global optimization problem on the entire domain into many smaller and simpler local optimization sub-problems. Table \ref{tab:ddm} summarizes the different domain decomposition methods (DDM) that have been developed for neural PDE solvers. They can be broadly classified into two categories. The first category is \emph{non-overlapping DDM}, where the domain is divided into non-overlapping subdomains. A separate neural network is trained on each subdomain, and continuity across subdomain interfaces is enforced using additional interface terms in the PINN loss function. DPINN \cite{dwivedi2021distributed}, XPINN \cite{jagtap2021extended}, cPINN \cite{jagtap2020conservative}, and hp-VPINN \cite{kharazmi2021hp} belong to this category. A key drawback of these approaches is inherent to their design in enforcing continuity across the subdomain interfaces. Since the interface conditions are only weakly constrained in the loss function, it can lead to artificial discontinuities at the subdomain interfaces. The additional interface terms not only introduce additional hyperparameters that need to be tuned to train the best model, they can also compete with the PDE losses. This contention can often slow convergence \cite{wang2022and}. Nevertheless, they are relatively simple to implement, and cPINN/XPINN have been parallelized to scale to multiple GPUs, leading to reductions in training times \cite{shukla2021parallel}. The second category is \emph{overlapping DDM}, which divides the domain into overlapping subdomains. In DeepDDM \cite{li2020deep} and D3M \cite{li2019d3m}, PINNs replace the subdomain solvers with variants of the classical Schwarz domain decomposition method \cite{lions1988schwarz}. FBPINN \cite{moseley2021finite} employs a separate input normalization in each subdomain and summation over all subdomain networks. \update{Continuity between interfaces is enforced via the construction of the PINN solution ansatz}\footnote{\update{A solution ansatz is a mathematical form or assumption about the solution to a PDE. It captures specific properties that the solution should satisfy, such as boundary conditions or initial conditions.}}. It can also be cast in the form of additive, multiplicative, and hybrid Schwarz methods \cite{dolean2022finite}. In contrast to the above approaches that require \emph{training separate neural PDE solvers} on non-overlapping or overlapping subdomains and resolving the interface between them, \mf \cite{wang2022mosaic} solves PDEs on larger domains using \emph{only inference}. An iterative algorithm inspired by the alternating Schwarz method updates the solution in subdomains using the pre-trained network inferences while maintaining the spatial regularity of the solution at subdomain boundaries. This eliminates the need to retrain the neural network for each new domain, making \mf highly reusable across different domain sizes and significantly reducing computational costs.

\begin{figure}[H]
    \captionsetup{skip=5pt}
    \centering
    \includegraphics[scale=0.34]{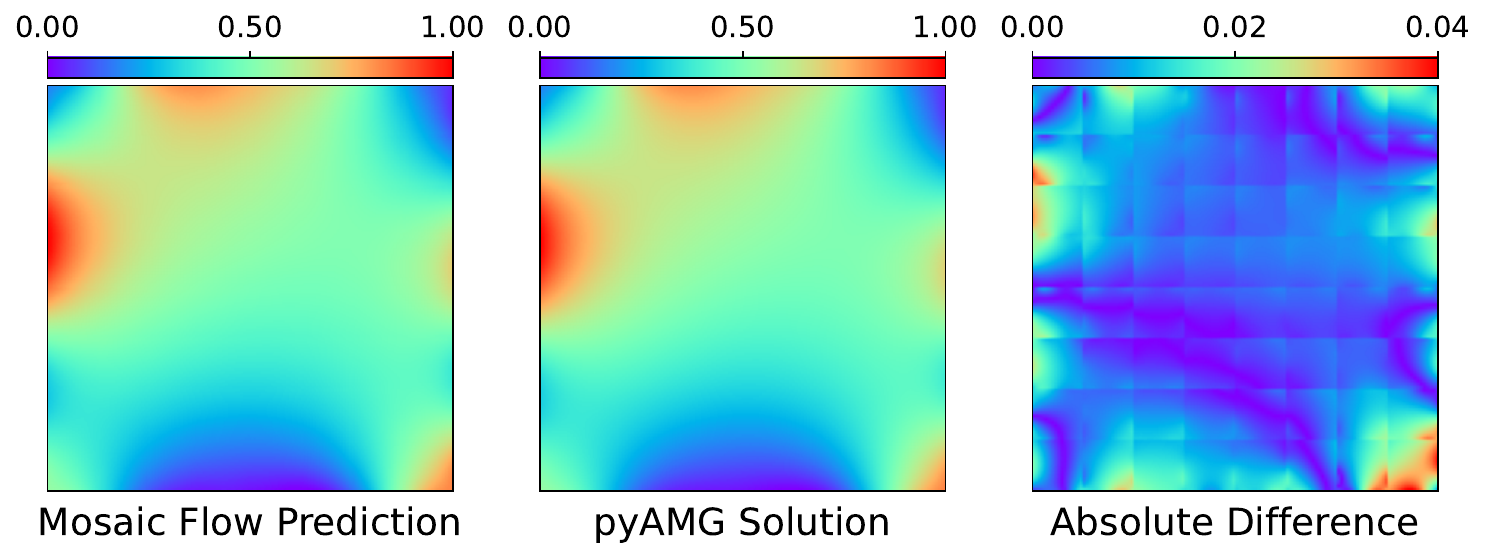}
    \caption{The leftmost sub-figure shows the solution using pyAMG to solve the Laplace equation on a $2\times2$ spatial domain with $128\times128$ resolution. The boundary condition is generated through a Gaussian process. The middle sub-figure shows the result of using distributed \mf predictor on the same domain. The rightmost sub-figure shows the absolute difference between the two.\label{fig:laplace-example}}
\end{figure}

\textbf{Contributions and Findings.} We propose an end-to-end parallelization pipeline for scaling \mf to large domains that encompasses both training and inference.
\begin{itemize}
    \item \textbf{Training on small domains.} We redesign the physics-informed neural PDE solver with focus on performance and scalability. The resulting network with an innovative input embedding and optimized architecture, combined with data-parallel training, reduces the training time for learning the Laplacian operator from hours to just $2$ minutes on 32 NVIDIA A30 GPUs.
    \item \textbf{Inference on large domains.} To enable scalable distributed inferences using the pre-trained neural PDE solvers on arbitrarily large domains, we propose a relaxation in synchronization. The relaxed distributed algorithm maintains accuracy as shown in Figure \ref{fig:laplace-example} while scaling up inferences to domains 4096 times larger for solving the Laplace equation. To the best of our knowledge, this is the largest domain solved in seconds using 32 GPUs, combining DDM with physics-informed neural PDE solvers as sub-domain solvers. In addition, we demonstrate strong and weak scaling up to 32 GPUs. 
\end{itemize}
\vspace{-0.25cm}
\section{Background} \label{sec:background}

This section begins with a brief introduction to the problem definition and physics-informed neural PDE solvers. We then delve into domain decomposition and elaborate on how \update{\mf} leverages this approach to implement large-scale physics-informed neural solvers.

\subsection{Problem Definition}

In this work, we develop SciML models to solve boundary value problems (BVP) of the form

\begin{equation}
\begin{aligned}
    D[u(\V{x})] = f(\V{x}),& \quad \V{x} \in \Omega \\
    B[u(\V{x})] = g(\V{x}),& \quad \V{x} \in \partial \Omega
\end{aligned}    
\end{equation}

The vectors $\V{x}$ are in the domain $\Omega$ or on the domain boundary $\partial\Omega$. The function $u$ is the solution of the differential equation. The differential operator is denoted by $D$, while $B$ is the boundary operator. The forcing function is $f$, and $g$ is the boundary function. A classic example of a BVP is the 2D Laplace equation with a Dirichlet boundary condition:

\begin{equation}
\begin{aligned}
    \Delta u(\V{x}) = 0,& \quad \V{x} \in \Omega \\
    u(\V{x}) = g(\V{x}),& \quad \V{x} \in \partial \Omega
\end{aligned}
\end{equation}

\update{where the vector $\V{x} = [x, y]$ and $\Delta = (\partial^2 / \partial x^2 + \partial^2 / \partial y^2)$ is the Laplacian operator \cite{evans10}.}

\subsection{Neural PDE Solvers}
Neural PDE solvers (or neural solvers for short) \cite{li2020fourier, kovachki2021neural, lu2021learning} are a type of model that learns to approximate the PDE \emph{solution operator} and solve various instances of a BVP with different boundary conditions. They are trained using a dataset that consists of solved boundary value problems for a specific PDE. A neural solver may take a discretized boundary function as input, denoted by $\hat{\vg} = \{g(\V{x}_{bc}^1), \dots, g(\V{x}_{bc}^N)\}$, where $\V{x}_{bc}^i$ are $N$ points sampled on the boundary. $\hat{\vg}$ specifies the particular instance of the BVP to solve. Therefore, an neural solver, represented by the function $\mathcal{N}(\V{x}, \hat{\vg}; \theta)$, approximates the solution $u(\vx)$ for the instance of the BVP determined by the boundary function $g$.

This study employs a special type of network called a \emph{physics-informed neural PDE solver} \cite{wang2021deeponet, raissi2019physics}. \update{While neural solvers trained on labeled input-output pairs can learn the solution operator of a PDE, their ability to generalize to out-of-distribution data, such as boundary or initial conditions outside the training set, significantly increases the demand on the training dataset size. In SciML, data is often scarce and computationally expensive to generate. To address this challenge, physics-informed neural solvers adopt a similar strategy to PINNs by incorporating an additional PDE loss as a form of regularization. This \emph{physics} loss effectively constrains the space of possible solutions, softly enforcing the PDE constraints. By incorporating domain knowledge into the training process, the model is more robust to noise and uncertainties present in the dataset. As an example, for the Laplace equation, the PDE loss for a batch of $n$ collocation points $\mX = \{\vx_1, \dots, \vx_n\} \subseteq \Omega$ is defined as}

\begin{equation}
\begin{aligned}
   \update{ \mathcal{L}_{pde}(\mX, \hat{\vg}; \theta) = \dfrac{1}{n}\sum_i^n (\Delta \mathcal{N}(\V{x}_i, \hat{\vg}; \theta)) ^ 2}
\end{aligned}
\end{equation}

Intuitively, as the loss function is minimized during training, the PDE residual will approach zero, $\Delta\mathcal{N}(\V{x}, \hat{\vg}; \theta) \rightarrow 0$, indicating that the network approximately satisfies the Laplace equation $\Delta\mathcal{N}(\V{x}, \hat{\vg}; \theta) \approx \Delta u(\V{x}) = 0$.

\subsection{Domain Decomposition}

Domain decomposition methods are widely used in solving boundary value problems \cite{balay2019petsc, hecht2012new, dolean2015introduction}. These methods partition the domain of a BVP into smaller subdomains, and then iteratively  combine solutions of the subdomains to develop the global solution. Domain decomposition methods enable scaling across multiple nodes, making them a powerful tool for scaling PDE solvers.

The classic example of domain decomposition is the alternating Schwarz method (ASM) \cite{schwarz1869ueber, gander2008schwarz}. ASM relies on overlapping subdomains to ensure continuity and information propagation between subdomains. While Schwarz methods are commonly used as preconditioners for Krylov methods \cite{dolean2016nonlinear}, in this work we use a variant of ASM as the solver.

Continuing with the Laplace example, the domain $\Omega$ can be partitioned into two subdomains $\Omega_1$ and $\Omega_2$, such that $\Omega_1 \cap \Omega_2 \not= \emptyset$. The subdomain interfaces are $\Gamma_1 = \partial \Omega_1 \cap \Omega_2$ and $\Gamma_2 = \Omega_1 \cap \partial \Omega_2$. To solve the global domain using ASM, the following routine is applied iteratively:

\begin{equation}
\begin{aligned}[c]
\Delta u_1^{n+1} &= 0 \text{ in } \Omega_1 \\
u_1^{n+1} &= u_2^n \text{ on } \Gamma_1
\end{aligned}
\qquad
\begin{aligned}[c]
\Delta u_2^{n+1} &= 0 \text{ in } \Omega_2 \\
u_2^{n+1} &= u_1^{n+1} \text{ on } \Gamma_2
\end{aligned}
\end{equation}

\update{The superscript denotes the iteration of the solution: $u^n_i$ is the solution of subdomain $i$ at iteration $n$. The process alternates between solving for $u_1$ and $u_2$. The solution for $u_1$ is used to set the interface condition on $\Gamma_2$. Then, with this condition, we solve for $u_2$. Subsequently, the solution for $u_2$ is used to set the interface condition on $\Gamma_1$. Schwarz proved the convergence of this iterative scheme for general elliptic PDEs \cite{schwarz1869ueber}.}
Lions proved that ASM can be used to solve systems with an arbitrary number of subdomains, and a parallel version of ASM exhibits similar convergence properties to the original Schwarz method \cite{lions1988schwarz}. However, it is worth noting that the convergence rate of Schwarz methods is signifiantly influenced by the mesh parameter and overlap. A system consisting of many subdomains with little overlap require more iterations to converge compared to a system with fewer subdomains and greater overlap. To address these issues, several improvements to the alternating Schwarz method have been proposed \cite{gander2006optimized, dubois2012optimized}. We leave exploring these improvements to future work.

\subsection{\mf}

\begin{figure*}[!ht]
	\centering
	\includegraphics[scale=0.6]{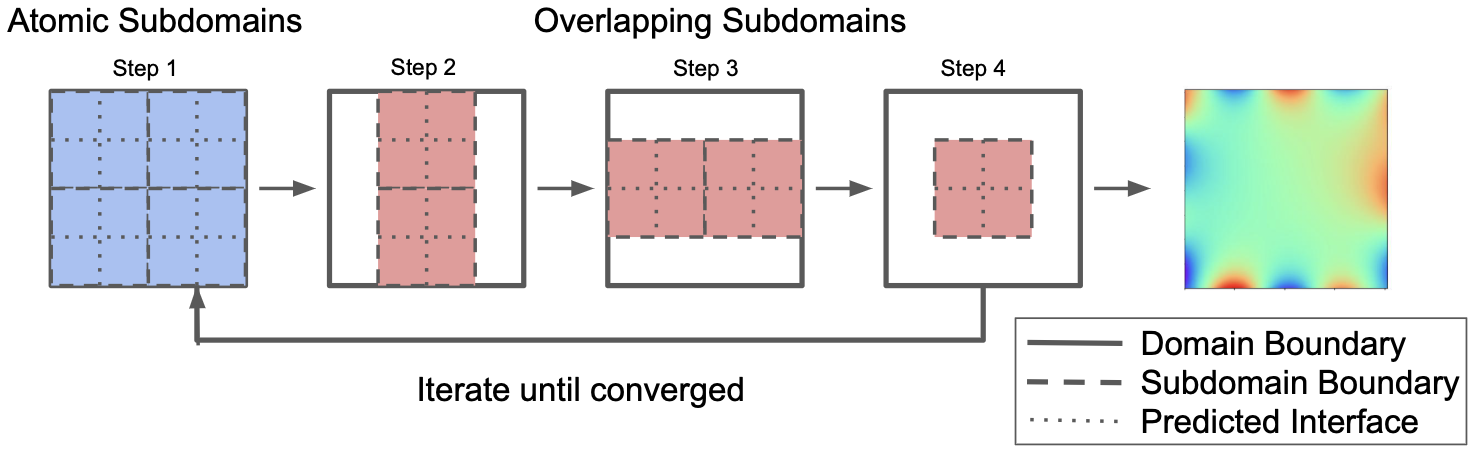}
\caption{\update{MFP predicts the solution in new domains} by combining SDNet predictions on \emph{atomic} and \emph{overlapping} subdomains. Unlike traditional numerical methods and PINNs, MF predictor only infers the solution on interfaces of subdomains rather than computing solutions for all grid or collocation points in the domain during each iteration.}
\label{fig:MFPredictor}
\end{figure*}

\mf \cite{wang2022mosaic} is a novel approach for solving BVPs on diverse domains with arbitrary boundary conditions. It consists of two primary components: 

\begin{enumerate}[leftmargin=*]
\item The subdomain solver (\textbf{SDNet}) is a \emph{physics-informed neural PDE solver} trained to solve a BVP on a small domain with arbitrary boundary conditions. Although this paper focuses on Dirichlet boundary conditions, SDNet can also be used with Neumann or Robin boundary conditions. SDNet's effectiveness arises from its ability to rapidly generate predictions for any point within the domain. Note that while the boundary function input to SDNet is discretized, the $xy$-coordinate can be in continuous space. This is unlike finite difference or finite elements methods, which require discretizing the interior of the domain \cite{Larrson2003pdenumerical}.

\item The \emph{\mf predictor} (\textbf{MFP}) illustrated in Figure \ref{fig:MFPredictor} is an iterative algorithm that leverages pre-trained SDNet's inferences to solve BVPs on large domains---much larger than those that can be solved with a SDNet directly. The iterative algorithm decomposes the domain into smaller \emph{atomic subdomains}, and updates the solution within each subdomain using SDNet's predictions. It also ensures the spatial regularity of the solution along the subdomain boundaries using \emph{overlapping subdomains}, inspired by the alternating Schwarz method. By utilizing SDNet as the sub-domain solver, MFP inherits its ability to make predictions for arbitrary points within the domain. This feature results in a significant performance advantage, as \mf can compute the solution for only a small fraction of grid points, specifically the interfaces of the subdomains, as opposed to all grid points in the entire domain, as done in classical ASM.
\end{enumerate}

\mf combines the efficiency of SDNet on small domains with the scalability of MFP on larger domains, enabling the efficient solution of complex BVPs.


\section{Neural PDE Solver Training} \label{sec:optimized-and-distributed-training}
SDNet is a neural PDE solver designed to approximate solutions to boundary value problems by taking a discretized boundary condition $\hat{\vg}$ and the coordinates of a point $\vx$ as inputs. $\mathcal{N}(\vx, \hat{\vg}; \theta) \approx u(\vx; g)$, where $u(\cdot; g)$ is the solution to the BVP determined by the boundary function $g$. By including the boundary condition in the input, the network can be used across multiple unseen instances of a BVP. However, this also results in a large input layer that, in combination with a PINN loss function, can make the network computationally expensive to train.

In the case of \mf, the training of physics-informed neural PDE solvers is restricted to a small domain for each PDE type. However, even on small domains (\eg spatial dimensions of $1 \times 1$), the training can take several hours (see Fig \ref{fig:gfnet-training}). To enable scalable training, performance-focused optimization and parallelization across multiple GPUs are crucial. By optimizing the training process, it becomes feasible to create a library of pre-trained SDNet models for different PDE types, facilitating the solution of complex multiphysics problems efficiently.

\subsection{SDNet Model Overview}

\begin{figure*}[!ht]
	\centering
	\includegraphics[scale=0.75, trim=1cm 1.75cm 2cm 0.5cm]{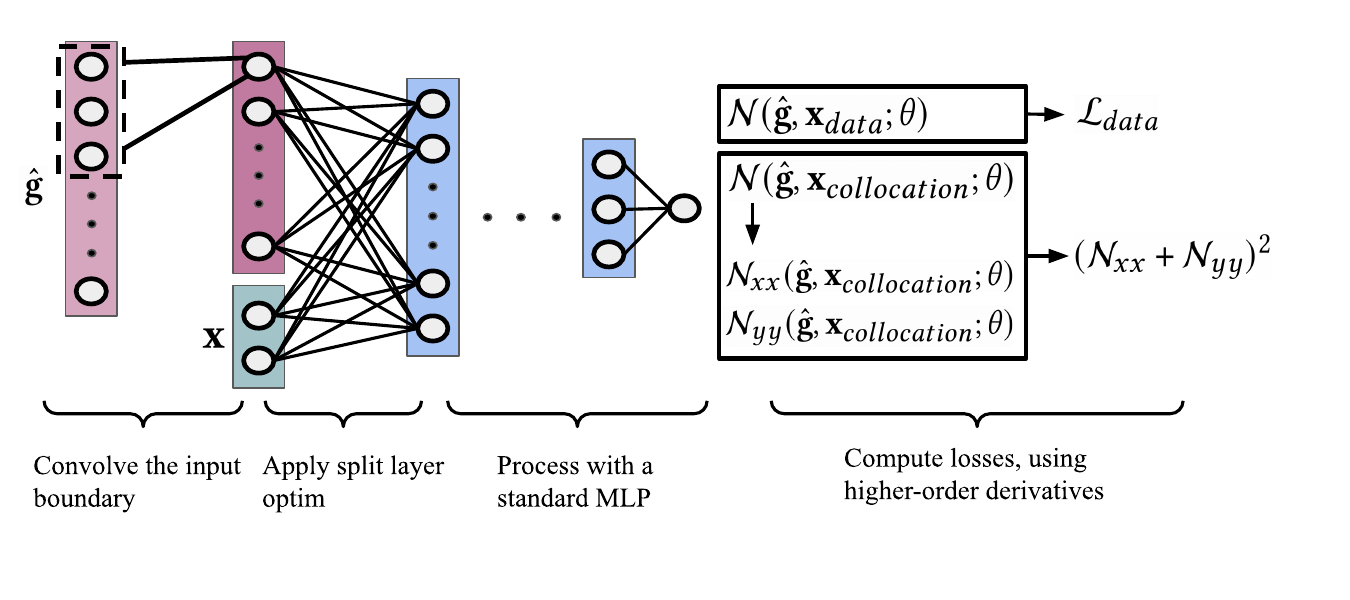}
	\caption{The architecture of the neural solver begins by convolving the input boundary condition, which results in a high-dimensional embedding. Next, the split layer optimization is applied to the batches of boundary conditions and $xy$-coordinates. The processed output is fed into a standard MLP, which approximates the solution of the PDE at the input points. Finally, the model computes higher-order derivatives to enforce the PDE constraints in the loss function. 
\label{fig:pinn-arch}}
\end{figure*}

\update{In general, our approach is agnostic of the choice of model for SDNet. For instance, one could use pure MLPs, a flavor of DeepONet \cite{lu2021learning}, or Fourier layers \cite{li2020fourier}.} The architecture of the neural solver used in this work, shown in Figure \ref{fig:pinn-arch}, is a variant of DeepONet that we inherit and improve. We first apply 1D convolutions to the input boundary conditions to create a high-dimensional embedding. The reason for using 1D convolutions is to take advantage of the inherent spatial structure of the boundary conditions, which can be seen as a 1D curve along the boundary of the domain. By convolving this signal, we capture local patterns and relevant features for predicting the solution within the domain. We anticipate this treatment of the boundaries will enhance convergence performance without affecting per-iteration performance, as convolutions are computationally efficient. We choose not to use Fourier layers due to the non-periodic nature of Dirichlet boundary conditions \cite{li2020fourier}. Although FNO can handle non-periodic boundaries, the combination of convolutions and fully-connected layers proves sufficient for capturing global phenomena.

Next, we apply the \emph{input-split} optimization discussed in Section \ref{optimized-input-embedding} to the high-dimensional boundary embedding coupled with the input $\vx$. The rest of the architecture is composed of a stack of linear layers, each followed by a nonlinear activation function. We use the \update{GELU activation function \cite{hendrycks2016gelu}} because PINN training tends to have better convergence properties when using a smooth activation function \cite{karniadakis2020convergence}.

\subsection{Optimized Input Embedding}
\label{optimized-input-embedding}
A common input embedding in physics-informed neural PDE solvers is to concatenate the spatial coordinates $\vx$ with the discretized boundary function $\hat{\vg}$ into a single input vector \cite{lu2021learning, wang2022mosaic}. However, this \emph{input-concat} approach is highly inefficient. 
For example, in a square 2D domain discretized into an $N\times N$ resolution, inferring the solution, $u(\vx)$ at a single point $\vx$ in the domain requires an input vector of dimension $4N+2$. The discretized boundary function $\hat{\vg}$ is a vector of dimension $4N$ and the additional $2$ dimensions are for the $xy$-coordinates.

When inferring the solution of $q$ points in a batch, the input becomes a $q \times (4N+2)$ matrix $\M{I}$:
\begin{equation}
      \M{I} = \begin{bmatrix} \hat{\vg}, \vx_1 \\ \hat{\vg}, \vx_2 \\ \vdots \\ \hat{\vg}, \vx_q\end{bmatrix} = \begin{bmatrix}
        \M{G}&  \M{X}
    \end{bmatrix}  \label{eq:inputMatrix}
\end{equation}

where the matrix $\M{G} \in \mathbb{R}^{q\times 4N}$ is formed by replicating the vector $\hat{\vg}$ for each point in the batch and the rows of $\M{X} \in \mathbb{R}^{q \times 2}$ are the coordinates of the $q$ points. 
Denoting the weights of the first layer of the neural network as $\M{W} \in \mathbb{R}^{d \times (4N+2) }$, the output of this layer is given by the matrix multiplication of $\M{I}$ with $\M{W}$, followed by the application of the activation function $\phi$. Mathematically, we can express this as:
\begin{equation}\label{eq:inputConcat}
    \M{U} = \phi(\M{I}\M{W}^\Tra)  \in \mathbb{R}^{q \times d}
\end{equation}

To reduce the computational cost and remove the redundancy in $\M{G}$ introduced by the \emph{input-concat} approach, we split weight matrix $\M{W} \in \mathbb{R}^{d \times (4N+2)}$ into two column blocks, denoted as
 $\M{W}_1 \in \mathbb{R}^{d\times 4N}$ and $\M{W}_2 \in \mathbb{R}^{d\times 2}$.
Using \cref{eq:inputMatrix}, we can rewrite \cref{eq:inputConcat} to arrive at our optimized approach:
\begin{align}
    \M{U} &= \phi(\begin{bmatrix}
        \M{G}&  \M{X}
    \end{bmatrix}\begin{bmatrix}
        \M{W}_1^\Tra \\[1ex] \M{W}_2^\Tra
    \end{bmatrix}) \\
      &= \phi(\hat{\vg}\M{W}_1^\Tra\oplus \M{X}\M{W}_2^\Tra )\label{eq:inputSplit}
\end{align}
where $\oplus$ is a broadcasted sum along the second axis of $\mX\mW_2^T$.
Note that the discretized boundary condition $\hat{\vg}$ is no longer replicated for each point in the input, but is computed only once and broadcasted along the batch dimension. This reduces the overall number of computations required by the network.
With \cref{eq:inputConcat}, the cost of the first layer is $O(qNd)$. In comparison, with our optimized \emph{input-split} approach in \cref{eq:inputSplit}, the cost is reduced to $O(Nd + qd)$. 
More importantly, this reduces the memory requirement of input tensor from $q(4N+2)$ words to $4N+2q$ words; when $q$ and $N$ are large, this saving can be substantial. 
The reduction in memory usage achieved by the optimized approach makes it possible to scale training to larger batch sizes. 

\subsection{Distributed Data Parallel Neural PDE solvers}

After optimizing the network architecture, we accelerate the training of neural PDE solvers with a physics-informed loss using data parallelism. Recall that when training a physics-informed model, we use a loss function with multiple terms: $\mathcal{L}(\theta) = \mathcal{L}_{data}(\theta) + \mathcal{L}_{pde}(\theta)$ where $\mathcal{L}_{data}(\theta)$ represents the \emph{data loss} function. The loss function that enforces the PDE constraints, $\mathcal{L}_{pde}(\theta)$ requires the computation of higher-order derivatives with respect to the model inputs. In the case of the Laplace equation, this involves calculating the second derivatives $\mathcal{N}_{xx}$ and $\mathcal{N}_{yy}$.
This results in a large autograd graph that consumes significant device memory. The size of this autograd graph limits the batch size on a single GPU, motivating the need to scale up to multiple GPUs to improve performance. It is worth noting that without the PDE loss, a purely data-driven model could be trained with a larger batch size on a single device. However, such a model may exhibit physical inaccuracies and require significantly larger dataset for training.

To efficiently train the network with multiple loss terms, we separate the data and collocation points into distinct forward passes. This approach simplifies the application of different losses to their respective coordinates, as the data loss can only be applied to points with known solutions. However, when using distributed data parallelism (DDP), it is important to preserve the standard semantics of stochastic gradient descent (SGD). \update{In data-parallel training, the model is replicated across different compute nodes, and local gradients are computed on each process. To synchronize gradients, an \emph{allreduce} \cite{chan2007collective} operation is commonly used, where the gradients are averaged across processes.} 
To maintain the correct semantics of SGD, we must be mindful of when gradient synchronization occurs. If synchronization occurs after both forward passes, it will compute a sum of averages rather than a true global average. Although this approach may yield satisfactory results in practice, it does not offer the same convergence guarantees.

\begin{algorithm}[H]
\centering
\begin{algorithmic}[1]
\State $\hat{\mG}$ is a batch of discretized boundary functions
\State $\mX_{data}$ is a batch of points with known solutions
\State $\mX_{collocation}$ is a batch of points with unknown solutions

\State \textbf{Step 1:} Solve Data Points
\State \hskip1em $\mP_{data}$ = $\mathcal{N}(\hat{\mG}, \mX_{data}; \theta)$
\State \hskip1em $\nabla_{data}$ = $\nabla \mathcal{L}_{data}(\mP_{data}; \theta)$

\State \textbf{Step 2:} Solve Collocation Points
\State \hskip1em $\mP_{collocation}$ = $\mathcal{N}(\hat{\mG}, \mX_{collocation}; \theta)$
\State \hskip1em $\nabla$ = $\nabla_{data}$ + $\nabla \mathcal{L}_{pde}(\mP_{collocation}; \theta)$
\State \textbf{Step 3:} \update{Perform \emph{allreduce} on gradient $\nabla$}
\end{algorithmic}
\caption{\update{SDNet Training Iteration}}
\label{alg:gfnet-train-iteration}
\end{algorithm}

To maintain consistency with SGD and ensure reliable convergence, we propose the method outlined in Algorithm \ref{alg:gfnet-train-iteration} for each training iteration. 
In step 1, the forward and backward passes are computed for the data points (lines 5-6) on each process without averaging gradients across processes. Then, in step 2, we apply the forward and backward passes for the collocation points (lines 8-9). The gradients for the collocation points are accumulated onto the gradients for the data points (line 9), and this sum is subsequently averaged across all processes using the allreduce operation in step 3. The averaged gradients are applied locally on each device, ensuring consistency.
\update{The proposed approach not only ensures accurate gradient accumulation but also offers the advantage of performing a single allreduce operation per training iteration, instead of two separate operations for data and collocation points. This optimization reduces communication overhead and enhances the scalability and efficiency of the training process.}

\section{Parallel and Distributed Inference}
The baseline MFP \cite{wang2022mosaic} has limited scalability, as we show in Section \ref{sec:results-mfp}. This constraint significantly hampers its capacity to solve BVPs on large domains. Our approach to addressing this limitation includes two strategies: increasing device-level parallelism and formulating the \emph{distributed MF predictor} algorithm for multi-GPU scaling. The algorithm is designed to harness the inherent strengths of the baseline MFP, while simultaneously extending its scope to BVPs on significantly larger domains.

\subsection{Batched Inference with Atomic Subdomains}\label{sec:seqInfOpt}
The baseline MFP adopts a sequential approach to solve subdomains, which ensures that all predictions are based on updated boundary conditions. Empirical evidence suggests that relaxing this can have a negative effect on prediction accuracy. However, by observing Figure \ref{fig:MFPredictor}, it becomes evident that atomic subdomains within each iteration of the algorithm do not overlap. This creates an opportunity for concurrently predicting these non-overlapping subdomains. To leverage this, we implement a batching technique that combines the atomic subdomains as input for SDNet inference. This effectively increases GPU occupancy by exploiting device-level parallelism as demonstrated in Section \ref{sec:batchedInference}.

\subsection{Domain Parallelization for Distributed Inference}\label{sec:distributedAlg}
The MFP takes the boundary conditions of subdomains as its input. 
During the development of the distributed algorithm, a key factor is both accurately and effectively managing updates to boundary conditions within overlapping regions.
The boundary information of the subdomains can be organized into a Cartesian grid.
In the example illustrated in \Cref{fig:MFPredictor}, the distance between neighboring grid points is $\frac{1}{2}m$, which is a tunable hyperparameter. Choosing a smaller distance allows for more subdomains to be placed in the domain, potentially resulting in more accurate results. However, this also leads to increased computation and communication costs, as shown in Section \ref{sec:DistMFCostAnalysis}. In this study, we choose a value of $\frac{1}{2}m$ because it is the largest distance we can use without significantly sacrificing accuracy. 

\begin{figure*}[!ht]
    \centering
    \input{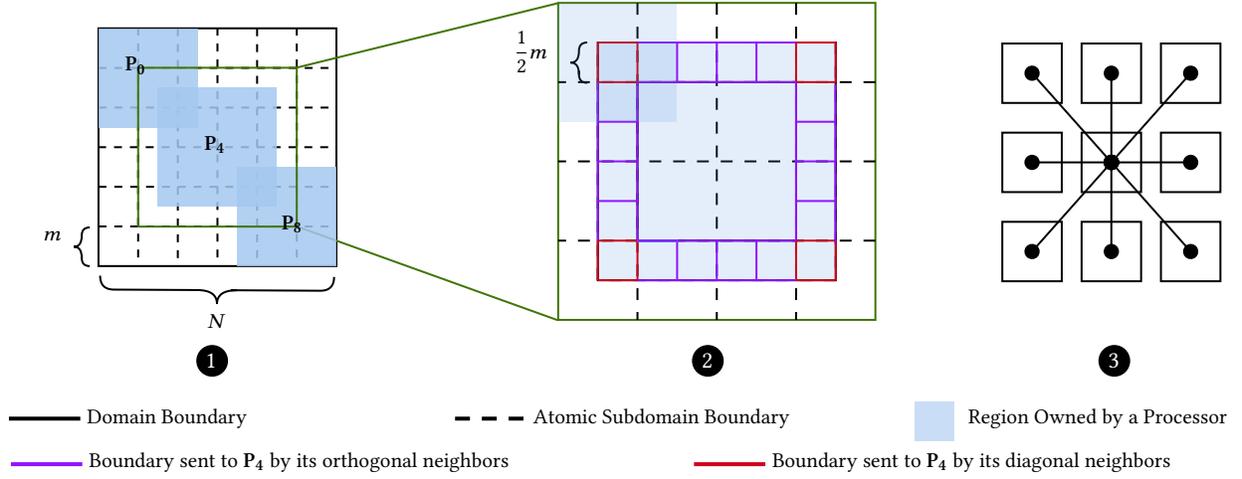}
    \caption[Figure 4]{\encircle{1} illustrates a $N\times N$ domain distributed among a $3\times 3$ processor grid and the placement of non-overlapping atomic subdomains on the entire domain. We omitted some of the processors in these figures to avoid cluttering. \encircle{2} zooms in on the region encompassing the subdomain owned by $P_4$. It highlights the boundaries sent to $P_4$ by its neighbors. The red lines indicate boundaries received by $P_4$ from its diagonal neighbors: $[P_0, P_2, P_6, P_8]$. The purple lines indicate boundaries received by $P_4$ from its orthogonal neighbors: $[P_1, P_3, P_5, P_7]$. \encircle{3} shows the stencil communication pattern for exchanging boundary information. For processors on the four boundaries, the communication group will not include all 9 processors. }
    \label{fig:distributionAndCommunication}
\end{figure*}

To design a parallel algorithm for the MFP, we first divide the global domain into a 2D grid, where each block is assigned to a processor. The processors are assigned to this 2D grid in a row-wise scan pattern. It is worth noting that a processor mapping strategy based on locality-preserving space-filling curves such as Morton order \cite{Morton1966} or Z-order could provide better load balancing and reduced data movement \cite{pekkila2022scalable}, although we leave this for future studies. We refer to the region owned by a processor as the \emph{processor subdomain}. In order to iteratively approach the final solution using Schwarz methods, neighboring processor subdomains need to communicate boundary information. 
To enable this exchange of information, we give each processor additional \emph{halo} boundary information from its neighboring processor subdomains. 
Figure \ref{fig:distributionAndCommunication} illustrates the distribution of processor subdomains and how the boundary information is exchanged between processors.

Our proposed distributed algorithm is outlined in \cref{alg:distmf}, where $t$, $\epsilon$, $\V{g}$, SDNet, and $\V{n}$ respectively denote the maximum number of iterations, convergence threshold, global boundary condition, the pre-trained SDNet model, and the neighbors of the current processor. We use the hat notation to denote a local variable (for example, $\hat{\V{g}}_0$ denotes the local part of $\V{g}_0$). The algorithm can be conceptually divided into two phases. The first phase is the iteration loop from \cref{line: loopStart} to \cref{line: loopEnd}. In each iteration, the boundary information of the local atomic subdomains is input to the pre-trained SDNet, which predicts the values \emph{only} along the center lines of these atomic subdomains (\cref{line: sdnet}). Since the center lines of one subdomain are the boundary of another, these predictions are subsequently input to SDNet for the next iteration. After the inference step, the boundary information in the region where processor subdomains overlaps is packed into a contiguous buffer and sent to the corresponding neighbors with \texttt{communicate\_new\_boundaries} in \cref{line:sendRecv}. \Cref{fig:distributionAndCommunication} provides a graphical illustration of which parts of the boundaries are being communicated. 
The second phase starts after $t$ iterations or upon reaching the convergence threshold. In this phase, each processor uses the most recent atomic subdomain boundaries as input for SDNet, to predict the values at every grid point within each atomic subdomain, forming the local solution $\hat{\M{S}}$ (\cref{line:localSolution}). Finally, an \texttt{all\_gather} is performed to collect the distributed subdomains. In the region where processor subdomains overlap, the final solution is obtained by computing the average of the predictions.

\begin{algorithm}
\centering
\begin{algorithmic}[1]
\Function{Predict}{$t$, $\epsilon$, SDNet, $\V{g}_0$, $\V{n}$}
\For{$i$ in 1:$t$}\label{line: loopStart}
    \State $\hat{\V{g}}_i$ = SDNet$(\hat{\V{g}}_{i-1})$ \label{line: sdnet}
    \State $\hat{\V{g}}_i$ = communicate\_new\_boundaries($\hat{\V{g}}_i$, $\V{n}$)  \label{line:sendRecv}
    \State $\delta = \|\hat{\V{g}}_i - \hat{\V{g}}_{i-1}\|/\|\hat{\V{g}}_{i-1}\|$
    \If{$\delta < \epsilon$}
        \State break
    \EndIf
\EndFor\label{line: loopEnd}
\State $\hat{\M{S}}$ = SDNet($\hat{\V{g}}_i$)\label{line:localSolution}
\State all\_gather($\hat{\M{S}}$)
\State S = combine all $\hat{\M{S}}$ and average over the overlapping regions
\State \Return $\M{S}$
\EndFunction
\end{algorithmic}
\caption{\update{Distributed Mosaic Flow Predictor}}
\label{alg:distmf}
\end{algorithm}

To design a scalable algorithm, we partially relax the order of subdomain updates in the baseline algorithm to balance accuracy requirements with communication efficiency. In the algorithm illustrated in \Cref{fig:MFPredictor}, as subdomains are solved by SDNet, the update to the boundary information inside the subdomain is applied immediately. However, when processors solve subdomains on the overlapped region, the update to the boundary information cannot be reflected in the neighboring processor until the communication step. In our parallel algorithm, we relax this principle by communicating only once per iteration. This relaxation in synchronization not only reduces the communication frequency but also makes the communication pattern agnostic to subdomain placement schemes.  It is worth highlighting that the baseline principle of immediate updates to boundary information still holds within individual processor subdomains. This relaxation is similar to the algorithm proposed by Lions \cite{lions1988schwarz}, which was proven to converge to the global solution. Empirical results in \Cref{sec:scalingPerformance} show that this modification does not prevent the distributed MFP from finding accurate solutions. 


\subsection{Cost Analysis}\label{sec:DistMFCostAnalysis}

We now analyze the costs associated with the distributed MF Predictor. Suppose the global domain has a resolution of $N\times N$ and is distributed across $P$ processors arranged in a $\sqrt{P}\times \sqrt{P}$ grid. The resolution of each subdomain is $m\times m$. 
As a result, each processor is assigned a subdomain consisting of $\frac{N}{m\sqrt{P}}\times \frac{N}{m\sqrt{P}}$ non-overlapping subdomains.
Assuming that the subdomain boundaries form a Cartesian grid with interval $\frac{m}{d}$, and the overlapping region is $\frac{2(d-1)}{d}m$ wide along each subdomain boundary, there are $\frac{(dN)^2}{m^2P}$ subdomains in each processor with all eight neighbors.
Using the alpha-beta model and removing the trailing terms, the communication cost of each processor is $C_{comm} = 8I\alpha + \frac{1}{\beta}I(16\frac{Nd}{\sqrt{P}})$, where $\alpha$ models the network latency and $\beta$ models the network bandwidth.

Since communication is performed in every iteration, both the bandwidth and latency cost scale linearly with the iteration count. As the communication is limited to each processor and its immediate neighbors, the latency cost is not influenced by $P$ or $N$. Bandwidth cost increases linearly with $\frac{N}{\sqrt{P}}$, which is the length of one side of each subdomain, and $d$, which controls how dense the subdomains are placed. 
Denoting the computation cost of making 1 SDNet inference as $c$, 
we can express the computation cost of each processor as $C_{comp} = c\frac{(dN)^2}{m^2P}$.
From this, we expect the computation cost to scale linearly with the number of processors. Note that we assume communication can be carried out simultaneously with all neighbors in our analysis, which may not always hold in practice.

\section{Results and Discussion}
\label{sec:results}

\update{We perform two sets of experiments to assess the performance of training and inference.} First, we evaluate SDNet training across multiple GPUs to analyze per-iteration performance and the impact of scaling on convergence. We present results on multiple GPU clusters, as detailed in \Cref{tab:cluster}, to gain a deeper understanding of the impact of optimizations discussed in Section \ref{sec:optimized-and-distributed-training}. Second, we evaluate the performance and scalability of distributed MFP on unseen domains that are significantly larger than the input seen by SDNet during training. \update{Ground truth data for both experiments is generated using the approach described in Section \ref{sec:data-generation}.}

\begin{table}[h]
 \centering
 	\begin{tabular}{|r|c|c|c|}
		\hline
	                       & V100      & A30     & A100      \\ \hline
        Architecture       & Volta     & Ampere  & Ampere     \\ 
        Peak FP32          & 14 TF     & 10.3 TF &  19.5 TF         \\ 
        GPUs/node          & 4         & 4       &  2         \\
        Nodes              & 13        & 14      &  4        \\ \hline
	    Memory          & 16GB      & 24GB    &  80GB    \\
                        &  (HBM2)   & (HBM2) & (HBM2e) \\
        Memory Bandwidth   & 900 GB/s  & 933 GB/s&  2 TB/s   \\ \hline
        Intra-node Interconnect & 32 GB/s   & 200 GB/s &  600 GB/s    \\	
                     & (PCIe)     & (NVLink)    & (NVLink)  \\ \hline	
        \update{Inter-node Interconnect} & \multicolumn{3}{c|}{100 Gbits/s}  \\ 
                    &   \multicolumn{3}{c|}{(ConnectX-5 Infiniband)}   \\ \hline	
	\end{tabular}
 \caption{GPU evaluation platforms and their specifications.}
  \label{tab:cluster}
\end{table}

\subsection{Data Generation} \label{sec:data-generation} 
\update{We generate two distinct datasets: one for training SDNet and another for evaluating the MFP. The training dataset consists of small domains of fixed size, while the test dataset includes larger domains of arbitrary sizes. To construct these datasets, we generate boundary conditions using Gaussian processes and follow a similar approach to the original \mf paper \cite{wang2022mosaic}. First, we use a Sobol Sequence \cite{SOBOL1998103} to sample the hyperparameters of an infinitely differentiable Gaussian kernel of a 1-dimensional Gaussian process. Then, from each Gaussian process, we draw a sample function (i.e., a 1-D curve). This function serves as the discretized boundary function $\hat{\vg}$ described in Section \ref{sec:background}. Each boundary value problem for the Laplace equation is solved using pyAMG. \cite{BeOlSc2022}.} 

\subsection{SDNet Training}

\paragraph{\textbf{Train Dataset.}} \update{We use the methodology described in Section \ref{sec:data-generation} to generate a dataset of $20,000$ boundary conditions for domains with a resolution of $32\times 32$ and spatial dimension of $0.5 \times 0.5$.} The pairs of boundary conditions and sample solutions form our training dataset. We use 90\% of this dataset for training and hold out the remaining 10\% as a validation set.

\paragraph{\textbf{Hyperparameters.}} \update{We perform hyperparameter tuning to determine the optimal values for several parameters, including the maximum learning rate, the fraction of iterations used for learning rate warmup, the learning rate schedule, the number of epochs, weight decay, and the number of points per subdomain. We do this tuning using a single GPU and select a sufficiently large batch size to ensure efficient GPU utilization.
The tuned hyperparameters we use are as follows: a maximum learning rate of $0.001$, using $0.1\%$ of iterations for learning rate warmup, using polynomial learning rate decay with the exponent set to one, training for $500$ epochs, and setting the coefficient for weight decay to zero.}

For experiments with varying GPU counts, we reuse the same hyperparameters from the single GPU case, with two modifications: 
\update{(a) We scale the maximum learning rate by the square root of the increase in batch size. (b) The fraction of iterations used for learning rate warmup is scaled linearly with the increase in batch size \cite{hsieh2020lamb, yin2021mlperf}.}
 
Finally, as we increase the number of GPUs, the number of points per batch can reach tens of thousands. We adopt the Lamb optimizer \cite{hsieh2020lamb}, which we find yields better convergence than AdamW \cite{loshchilov2018decoupled} when scaling to larger batch sizes and multiple GPUs. Specifically, we utilize the implementation of FusedLAMB from Nvidia Apex.

\begin{figure*}[!ht]
    \begin{subfigure}{0.45\textwidth}
    \centering
	\includegraphics[scale=0.35]{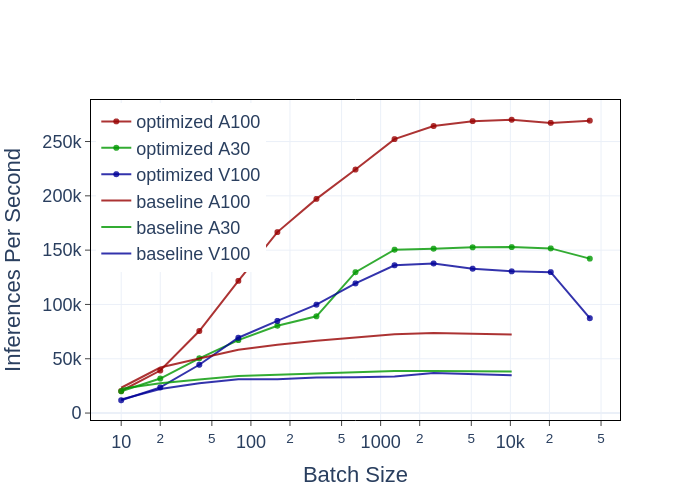}
    \caption{SDNet Inference vs. Batch Size}
    \end{subfigure}%
    \begin{subfigure}{0.45\textwidth}
    \centering
    \includegraphics[scale=0.35]{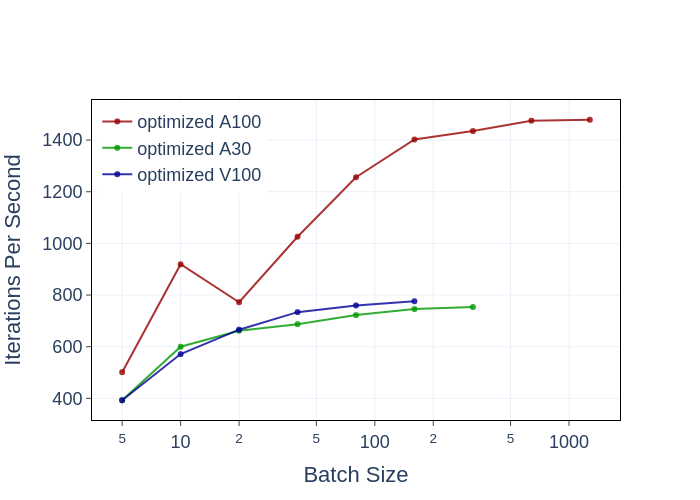}  
    \caption{SDNet Training Performance vs. Batch Size}
    \end{subfigure}%
 
	\caption{SDNet inference and training performance with varying batch sizes. \emph{optimized} model utilizes the split-layer optimization, while the \emph{baseline} model is a standard neural PDE solver. \update{Each point is the average of 30 trials. The variance is near zero in every case.} This plot shows both how the split-layer optimization improves performance, and enables scaling to larger batch sizes. For instance, the baseline models reach memory limits at a batch size of $10,000$ points, while the optimized models can scale to larger batch sizes, processing up to $50,000$ points during inference.
 }
	\label{fig:gfnet-inference-perf}
\end{figure*}

\paragraph{\textbf{Training Methodology.}} To train the SDNet, we employ a loss function with two terms: a data loss and a PDE loss. The data loss is a mean squared error using the pyAMG solution as the ground truth. The PDE loss is the PDE residual applied at the collocation points. It requires computing higher order derivatives with respect to the model inputs. Despite the relatively small size of our models compared to state-of-the-art vision and language models, the autograd graph generated during training consumes a significant amount of device memory. This memory constraint severely limits the batch size that can be used on a single GPU, which motivates the distributed data parallel approach to training. As seen in Figure \ref{fig:gfnet-inference-perf}, we can scale inference to process hundreds of thousands of subdomains at a time, but merely hundreds during training. Even for a relatively simple PDE like Laplace, a single model update requires three backward passes: (a) a backward pass to compute the derivatives w.r.t $x$ and $y$, (b) a second backward pass to compute the \textit{second} derivatives w.r.t. $x$ and $y$ and (c) a final backward pass, through the prior two gradient computations. \update{We measure the maximum memory allocated during the forward and backward passes of the model. The results, presented in Table \ref{tab:memory-consumption}, highlight the difference in memory usage with and without the PDE loss.  The inclusion of the PDE loss leads to a significant increase in memory consumption, primarily attributed to the storage of intermediate activations in the autograd graph.}

\begin{table}[h]
    \centering
    \begin{tabular}{c|c|c}
        \# Domains & No PDE Loss & With PDE Loss \\
        \hline
        5 & 0.05 GB & 0.503 G \\
        320 & 2.77 GB & 15.11 GB \\
        640 & 5.54 GB & OOM \\
    \end{tabular}
    \caption{\update{Memory allocated during the forward pass, loss computation, and backward pass on a single V100 GPU with and without PDE loss. OOM indicates ``out of memory''.}}
    \label{tab:memory-consumption}
\end{table}

\begin{figure*}[!ht]
\begin{subfigure}{0.33\textwidth}
    \centering
    \caption{Epoch Counter vs. MSE}
    \includegraphics[scale=0.31]{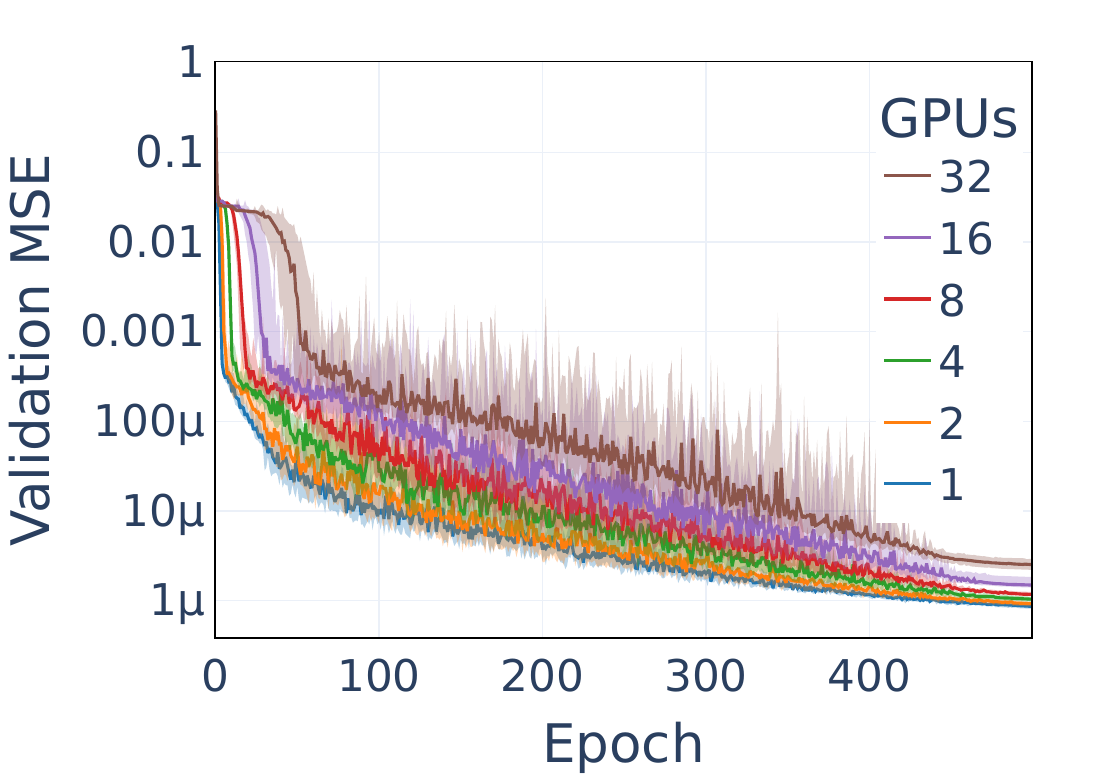}
\end{subfigure}
\begin{subfigure}{0.33\textwidth}
    \centering
    \caption{Runtime vs. MSE}
    \includegraphics[scale=0.31]{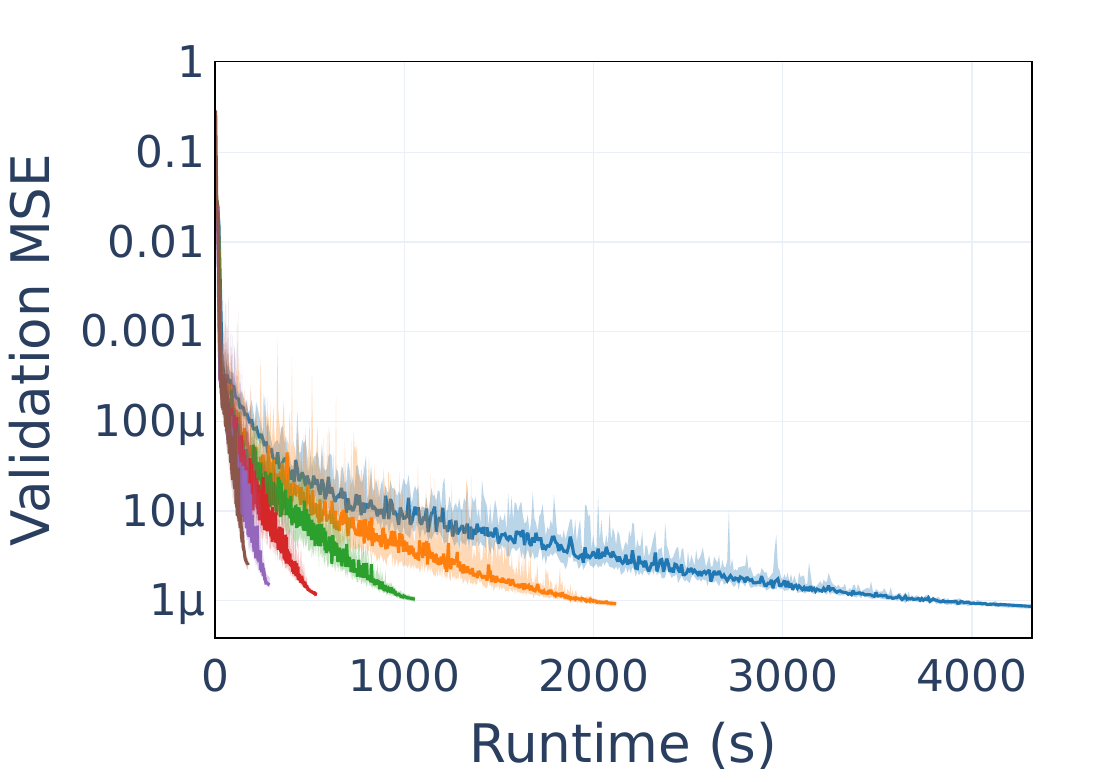}
\end{subfigure}
\begin{subfigure}{0.33\textwidth}
    \centering
    \caption{Runtime to MSE $2.5\times 10^{-6}$}
    \includegraphics[scale=0.31]{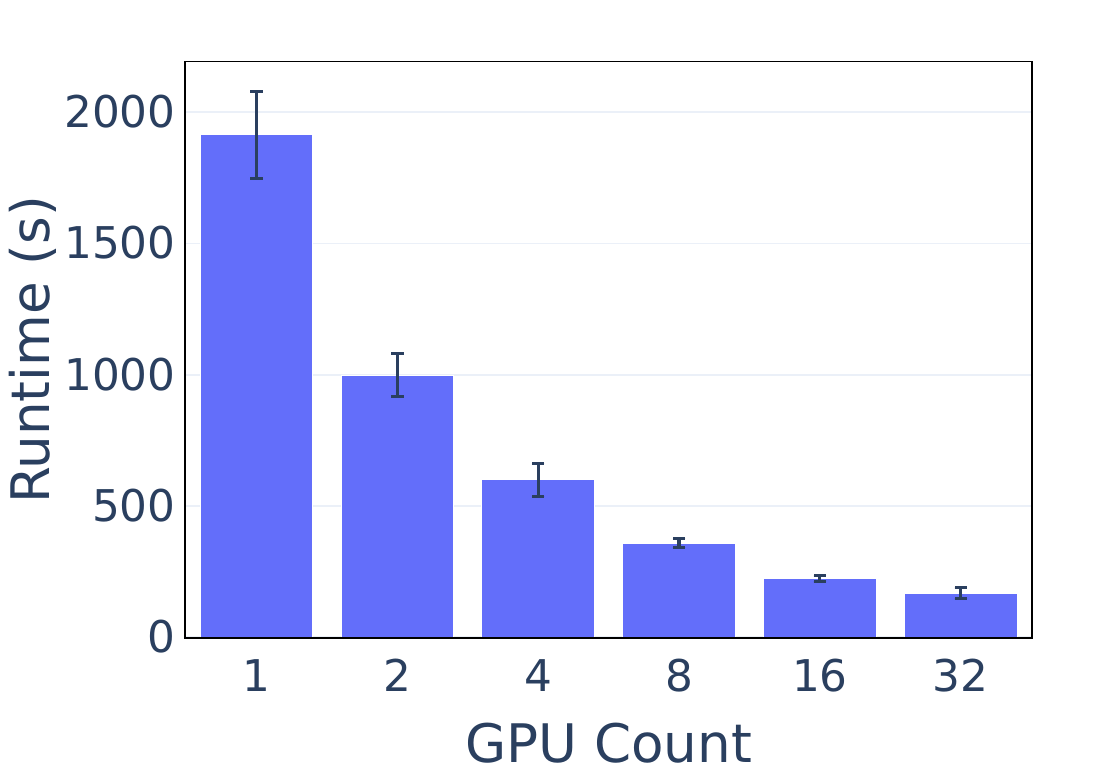}
\end{subfigure}
\caption{\update{ SDNet multi-GPU performance and impact on convergence.
\textbf{(a)} shows the MSE of the validation set as a function of the epoch count.
\textbf{(b)} illustrates the runtime improvements with increasing A30 GPUs.
Both (a) and (b) report the median validation MSE across 10 models initialized with different random seeds. The bands represent the $95\%$ confidence interval of the median \cite{torsten2015benchmark, leboudec2010performance}. Note that (a) and (b) are plotted on a $log_{10}$ scale and all models achieve final MSEs within $1.5\times 10^{-6}$ of the single GPU case.
\textbf{(c)} shows the average time, across 10 trials, taken by each model to reach an MSE of $2.5\times 10^{-6}$, which corresponds to the mean MSE of the final epoch with 32 A30 GPUs. The bands in (c) represent the standard deviation.}}

	\label{fig:gfnet-training}
\end{figure*}

We implement data parallel training using PyTorch Distributed. A key advantage of PyTorch's implementation of DDP training is the ability to  overlap communication with the current backward pass \cite{soumith2020pytorch}. This is unlike other frameworks, like Horovod, which overlap communication with the following forward pass. It is important to ensure that communication overhead does not dominate the overall execution time. Since our models are relatively small, the forward passes are typically inexpensive. Therefore, overlapping communication with the current backward pass improves the efficiency of training our models.

\paragraph{\textbf{Training Performance}}

We implemented several optimizations that result in much faster training compared to a baseline neural solver. First, we implemented the split layer, which significantly reduces redundant computation in the first layer of the network. This optimization is also important for the performance of model inference in the MFP, as seen in Figure \ref{fig:gfnet-inference-perf}. Second, we apply a series of 1-dimensional convolutions to the input boundary conditions, which form a smooth curve. Convolutions are cheap to compute, so this optimization has essentially no effect on the per-iteration performance of the MFP, but improves the convergence rate of the SDNet. Finally, we scale model training across multiple GPUs.  


Figure \ref{fig:gfnet-training} shows the performance and accuracy of SDNet when scaling the number of GPUs. \update{Although, we observe a slight negative impact on the validation MSE, all models achieve final MSEs within $1.5\times 10^{-6}$ of the model trained on a single A30 GPU. Notably, the model trained on one GPU takes over 30 minutes to reach an MSE of $2.5\times 10^{-6}$, while 32 GPUs reduces the training time to just \emph{two minutes} to reach the same MSE, resulting in a speedup of $12\times$.}

\begin{figure}[H]
    \centering
    \includegraphics[scale=0.28]{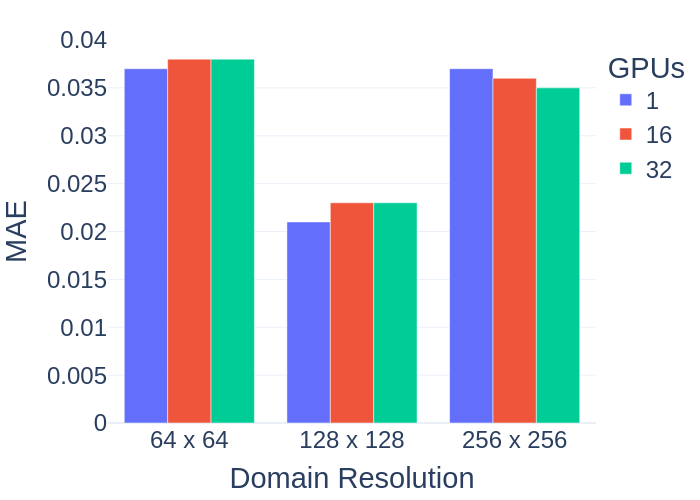}
    \caption{\update{MAE of the MFP, using models trained with varying GPU count. The discretized boundary function for each domain is $\hat{\vg}(x) = sin(2\pi x)$. This illustrates that the small changes in MSE seen in Figure \ref{fig:gfnet-training} have little affect on the MFP, which makes prediction of similar quality with each model.}}
    \label{fig:gfnet-eval}
\end{figure}

To compare the effectiveness of the SDNet models as sub-domain solvers for MFP, we additionally evaluate each SDNet on test problems of different sizes, as shown in Figure \ref{fig:gfnet-eval}. Despite the slight variations in the validation set's MSE (see Figure \ref{fig:gfnet-training}), we observe consistent MAE across all models. This indicates that all models exhibit comparable accuracy and are equally reliable as sub-domain solvers for MFP.

\subsection{MF Predictor Performance}
\label{sec:results-mfp}
We implement the distributed MFP in Python. For GPU-to-GPU communication, we use \texttt{mpi4py} \cite{dalcin2021mpi4py}, which is built with a CUDA-aware MPI library to enhance communication performance. To generate boundary conditions and ground truth solutions of the Laplace equation on larger domains, we use the method described in \Cref{sec:data-generation}. We evaluate both batched inference for device-level parallelism and distributed inference for node-level parallelism.

\begin{figure}[h]
\captionsetup{skip=5pt}
    \centering
    \includegraphics[scale=0.5]{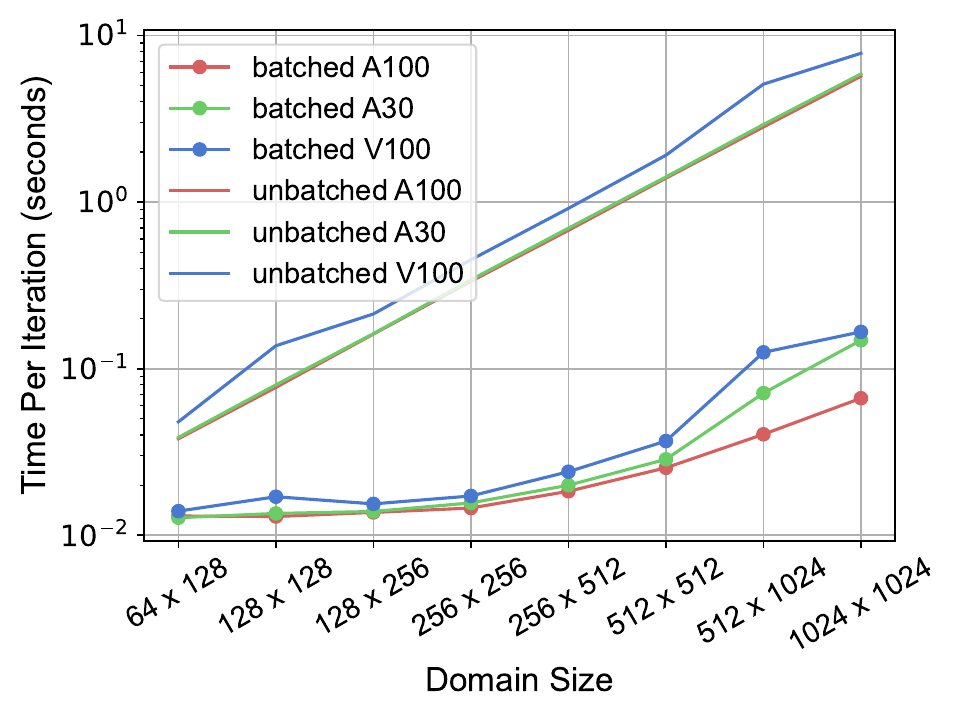}
    \caption{Performance of batched vs. unbatched atomic subdomains on a single GPU with increasing domain sizes. Time per iteration is calculated by averaging over 100 iterations.
    }
    \label{fig:batchedVunbatched}
\end{figure}

\paragraph{\textbf{Batched Inference.}}\label{sec:batchedInference}
In this experiment, we assess the performance improvement achieved by batching the atomic subdomains during each iteration of the MFP (as discussed in Section \ref{sec:batchedInference}). We compare this \emph{batched} approach to the original \emph{unbatched} algorithm, which predicts one subdomain at a time using SDNet. The results in Figure \ref{fig:batchedVunbatched}, shows the impact of batching when scaling the domain size from $1\times2$ to $16\times16$ (i.e., resolutions from $64 \times128 $ to $1024 \times 1024$). In the unbatched approach, time increases linearly with the domain size. However, with batching subdomains, we observe a significant improvement in GPU utilization, resulting in about 50\% of the peak performance. Note that since atomic subdomain inferences are independent, batching improves performance by up to 100$\times$ without sacrificing accuracy.  

\begin{figure}[htbp]
    \centering
    \begin{subfigure}[t]{\linewidth}
        \centering
        \includegraphics[scale=0.5]{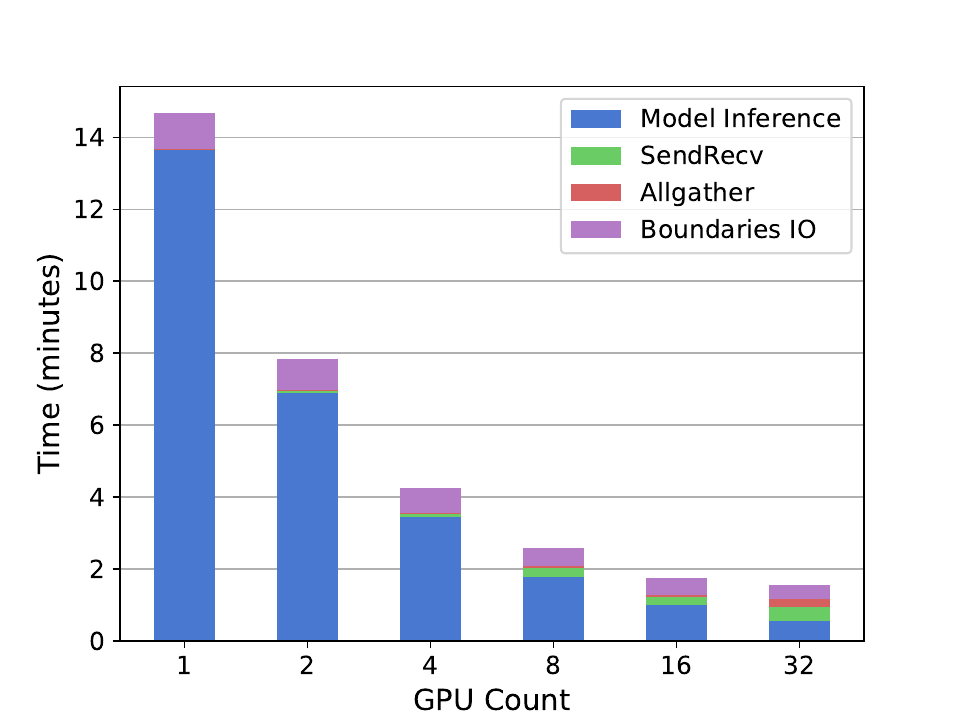}
        \caption{Strong scaling over a 2048$\times$2048 resolution domain.}
        \label{fig:strongScaling}
    \end{subfigure}
    \begin{subfigure}[t]{\linewidth}
        \centering
        \includegraphics[scale=0.5]{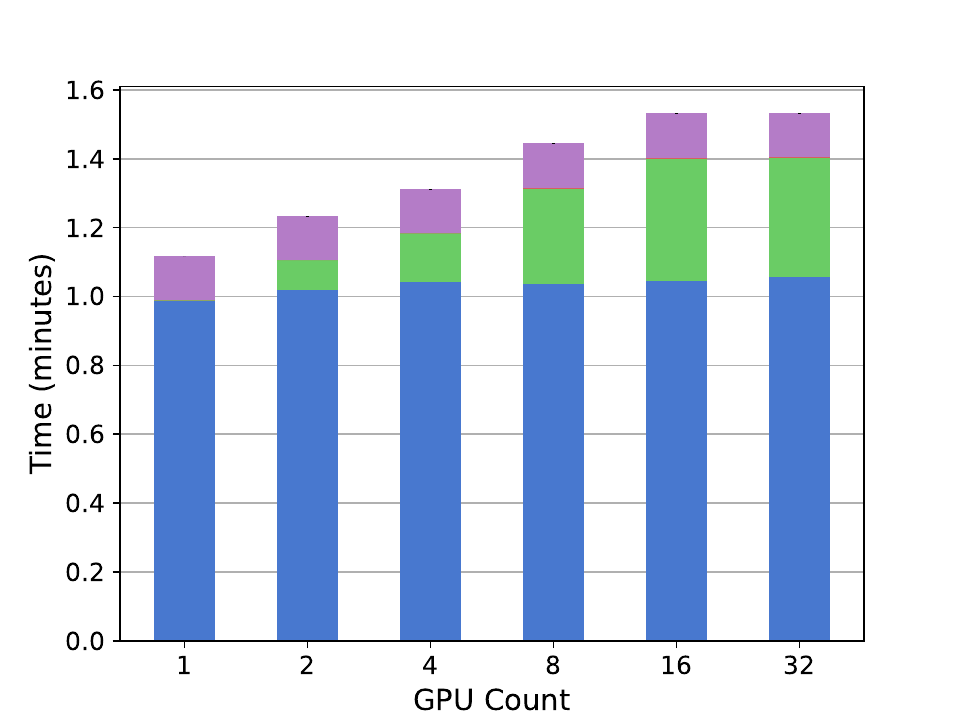}
        \caption{Weak scaling for 2000 iterations where each GPU owns a 1024$\times$512 resolution subdomain.}
        \label{fig:weakScaling}
    \end{subfigure}
    \caption{\update{Strong and weak scaling of MFP. We average the results across 5 trials. ``Boundaries IO" refers to reading subdomain boundaries for SDNet and updating them with the prediction from SDNet. }}
\end{figure}

\paragraph{\textbf{Distributed Inference.}}\label{sec:scalingPerformance}
We conduct both strong and weak scaling studies to evaluate MFP on multiple GPUs. In the strong scaling experiments, we consider a BVP for the Laplace equation with a spatial domain size of $32\times32$ ($2048\times 2048$ resolution). This domain is divided into $4096$ atomic subdomains where each subdomain is of size $0.5\times0.5$. The global boundary condition is generated using the same process described in \Cref{sec:data-generation}. The MFP terminates when the MAE of the solution drops below 0.05. The results, shown in \Cref{fig:strongScaling}, demonstrate a clear trend of decreasing computation time and an increasing percentage of communication time as we scale from 1 to 32 GPUs. The total runtime reduces from approximately 15 minutes ($\sim$ 880 seconds) to less than 2 minutes ($\sim$ 90 seconds), resulting in a speedup of almost $10\times$ on 32 A30 GPUs.

As discussed in \Cref{sec:distributedAlg}, updates in the overlapping regions along the borders of processor subdomains are not immediately reflected since the data is distributed. Therefore, as we decompose a domain into more (and smaller) processor subdomains, a larger percentage of the boundary information becomes stale. This can lead to a degradation of the convergence rate of the distributed MFP algorithm. In the strong scaling experiment, we investigate the impact of the distributed algorithm on the convergence rate. We record the number of iterations required to reach an MAE of 0.05 and present the results in Table \ref{tab:strongScalingIter}. As the number of processors increases, we observe a slight increase in the number of iterations required to reach the specified MAE. However, note that the benefits of parallelization and the reduction in computation time outweigh the slight increase in the number of iterations, leading to improved overall performance.

\begin{table}[htbp]
    \centering
    \begin{tabular}{c|c|c|c|c|c|c}
        GPU Count  & 1 & 2 & 4 & 8 & 16 & 32 \\\hline
        Iterations & 3200 & 3250 & 3250 & 3300 & 3400 & 3500
    \end{tabular} 
    \caption{\update{The number of iterations required to achieve a MAE of 0.05 for different GPU counts. The corresponding runtimes are shown in Figure \ref{fig:strongScaling}.}}
    \label{tab:strongScalingIter}
\end{table}

We also perform a weak scaling experiment with an increasing number of processors while keeping the spatial size of each processor subdomain fixed at $16\times8$ ($1024\times 512$ resolution), this result is shown in \Cref{fig:weakScaling}. Computation scales well, as the only additional computation cost is to average across regions where processor domains overlap. However, the communication scaling is less optimal. We see around $4\times$ increase going from 2 to 8 GPUs, which then plateaus. This increase is likely due to high latency cost as the number of messages sent by each processor increases with an increasing number of neighbors from 2 to 8 GPUs. We don't see a noticeable improvement in performance with CUDA-aware MPI compared to standard MPI, potentially due to the small buffer sizes of \texttt{send/recv} communication where latency dominates the overall communication performance \cite{dalcin2021mpi4py}. The increased latency cost is further exacerbated by \texttt{mpi4py}, which serializes Pytorch tensors before communication. Techniques that leverage direct GPU-to-GPU communication through NVSHMEM \cite{Ismayilov2023} are potential alternatives to reduce this communication overhead. 

\textbf{Open problems.}
\emph{Systems challenges --} One approach to addressing the latency overhead is to convert a latency-bound algorithm to a bandwidth-bound algorithm. This can be achieved by reducing the communication frequency. In the current implementation, each processor exchanges boundary information with its neighbors during every iteration. However, communicating less frequently introduces a trade-off with redundant computation. Given that compute scales significantly better than communication (both bandwidth and latency), \emph{communication-avoiding algorithms} are worth exploring. Nonetheless, there is a communication lower bound that cannot be avoided, in which case, overlapping communication with computation can further push the scaling ceiling. It is worth noting that \emph{communication-overlapping algorithms} have been well-studied in the context of numerical simulations \cite{pekkila2022scalable, wang2020pencil, Ismayilov2023}. However, neural PDE solvers can be significantly faster than numerical solvers. In contrast to large language models, current neural models for approximating PDEs are notably smaller. Additionally, batched inference only requires a forward pass (no expensive higher-order gradient computation). Consequently, communication becomes the \emph{bottleneck} for scaling even on smaller GPU clusters. Studying the trade-offs of communication-avoiding and communication-overlapping algorithms in the context of distributed neural PDE solvers remains a promising direction for future research.

\emph{Algorithmic challenges --} For BVPs, where you are interested in finding a solution that satisfies specific boundary conditions within a domain, information needs to be exchanged across the entire domain. For this reason, one-level Schwarz methods require a global coarse grid correction to scale to a large number of subdomains for solving  BVPs \cite{dubois2009schwarzweak}. FBPINN extended to multiple levels of overlapping domain decomposition demonstrates improved accuracy, specifically for large number of subdomains, implying that coarse levels are necessary for efficient global information propagation in large domains  \cite{dolean2023multilevel}. However, for time-dependent problems, where the solution evolves over time, information typically only needs to be exchanged between neighboring subdomains. As time advances, information is propagated across the domain as adjacent subdomains continually share their updated information. We hypothesize that distributed \mf coupled with one-level Schwarz is optimal for exploring neural domain decomposition methods to solve time-dependent PDEs \cite{takamoto2022pdebench, hassan2023bubbleml}. 

\section{Conclusions}

The hybrid parallelization scheme presented in this paper shows promise in scaling physics-informed neural PDE solvers to large domains using a combination of data parallel training and domain parallelization. The SDNets can be trained in minutes, allowing for the creation of a library of models for different PDEs. The MF Predictor demonstrated accuracy when scaling up to 32 GPUs.
Overall, this work opens up avenues for future research in the field of physics-informed machine learning. There is still room for improvement by exploring other domain decomposition methods and improved Schwarz methods, such as using a coarse correction \cite{dubois2009schwarzweak} or Optimized Schwarz methods \cite{gander2006optimized}, and extending DDM for time-dependent neural PDE solvers.

\begin{acks}
This work is supported by the National Science Foundation under the award number 2211908. We gratefully acknowledge the GPU computing resources provided on HPC3, a high-performance computing cluster operated by the Research Cyberinfrastructure Center at the University of California, Irvine. We specifically thank Hengjie Wang at Modular for helpful discussions on Mosaic Flow.
\end{acks}
\bibliographystyle{ACM-Reference-Format}
\bibliography{main}


\end{document}